\documentclass{Interspeech2024}

\interspeechcameraready

\title{Speed of Light Exact Greedy Decoding for RNN-T Speech Recognition Models on GPU}

\name[affiliation={1}]{Daniel}{Galvez}
\name[affiliation={1}]{Vladimir}{Bataev}
\name[affiliation={1}]{Hainan}{Xu}
\name[affiliation={1}]{Tim}{Kaldewey}

\address{
  $^1$NVIDIA, USA}
\email{dgalvez@nvidia.com, vbataev@nvidia.com, hainanx@nvidia.com, tkaldewey@nvidia.com}


\usepackage{algorithm}
\usepackage{algpseudocode}
\usepackage[export]{adjustbox} 

\begin{document}

\maketitle

\begin{abstract}

The vast majority of inference time for RNN Transducer (RNN-T) models today is spent on decoding. Current state-of-the-art RNN-T decoding implementations leave the GPU idle ~80\% of the time. Leveraging a new CUDA 12.4 feature, CUDA graph conditional nodes, we present an exact GPU-based implementation of greedy decoding for RNN-T models that eliminates this idle time. Our optimizations speed up a 1.1 billion parameter RNN-T model end-to-end by a factor of 2.5x. This technique can applied to
the ``label looping" alternative greedy decoding algorithm as well, achieving 1.7x and 1.4x end-to-end speedups when applied to 1.1 billion parameter RNN-T and Token and Duration Transducer models respectively.
This work enables a 1.1 billion parameter RNN-T model to run only 16\% slower than a similarly sized CTC model, contradicting the common belief that RNN-T models are not suitable for high throughput inference. The implementation is available in NVIDIA NeMo.

\end{abstract}

\section{Introduction}

RNN-T \cite{rnnt} is a model architecture favored for its high accuracy. 
It is commonly used for on-device speech recognition in production. 
However, because of poor economics stemming from low throughput of existing implementations, it does not see much production usage in GPU-based, server-side speech recognition. CTC \cite{ctc} models are generally preferred, because their non-autoregressive greedy and beam search decoders are straightforward to implement efficiently. The prevailing wisdom among speech recognition practitioners is that the autoregressive nature of RNN-T inference makes it unsuitable for running on GPUs, but the exact reason has never been explored in depth. We demonstrate in this paper what the precise problem is, and solve it in an exact way. By ``exact", we mean that our implementation computes exactly the same outputs as the baseline implementation for the same inputs.

For reference throughout the paper, we include Python pseudocode for the RNN-T greedy decoding algorithm in Algorithm \ref{rnntgreedydecoding}. This algorithm assumes the prediction network is a single layer RNN for simplicity, but any kind of prediction network can be used with our implementation. Every named variable in this algorithm is a torch tensor on the GPU. Some are annotated with a data type and a dimension. ``x:float[B,T,F]" means that x is a 3D floating point tensor with dimensions B(atch), T(ime), and F(eature).

\begin{figure}
    \centering
    \includegraphics[width=1.0\linewidth]{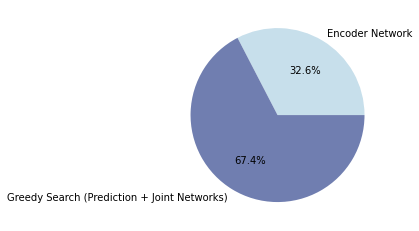}
    \caption{Time spent in the greedy decoder vs. the encoder in NVIDIA NeMo's Parakeet RNN-T 1.1b model to transcribe Librispeech Test Other, a 5.1 hour dataset, at batch size 32. The total runtime is 44.28 seconds on an A100-80GiB GPU.}
    \label{fig:encoder-vs-decoder}
\end{figure}

In a state of the art RNN-T model, Parakeet RNN-T 1.1B --at the time of writing third on the Open ASR Leaderboard \cite{openasrleaderboard}, achieving a word error rate of 7.04\%-- spends 67\% of its inference runtime on greedy decoding, while only 33\% is spent on the encoder at batch size 32, c.f., Figure \ref{fig:encoder-vs-decoder}. 
The encoder has 1.1 billion parameters, while the prediction and joint networks combined contain two orders of magnitude fewer parameters, ~8.9 million. In other words, 67\% of the runtime is spent on less than 1\% of the model's parameters.

\begin{figure*}[t]
    \centering
    \includegraphics[width=1\linewidth]{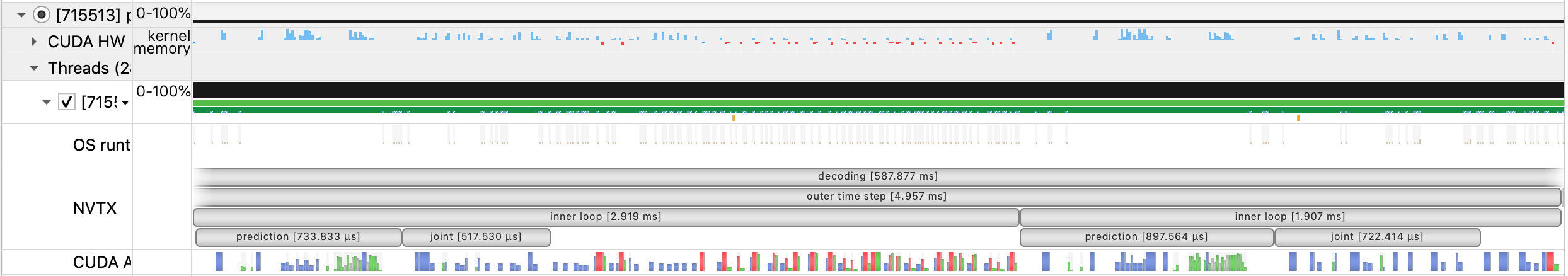}
    \caption{Nsight Systems profile of one ``outer time step" iteration of the loop in line \ref{alg:outerloop} of Algorithm \ref{rnntgreedydecoding}. Notice that the ``CUDA HW" row above has several blank regions. This indicates that the GPU is idle during those times.}
    \label{fig:trace}
\end{figure*}

Figure \ref{fig:trace} illustrates the reason the greedy decoder is significantly slower than the encoder. 
It depicts the time breakdown of tasks being done by line \ref{alg:outerloop} of Algorithm \ref{rnntgreedydecoding}, or one iteration of the ``outer time step" that goes over every time step of the ``encoder output". The ``inner loop" that keeps going until either max\_symbols have been produced or every element of the batch has produced blanks spends a significant amount of time not executing work on the GPU. The ``CUDA HW" row, which is light blue when the GPU is running a kernel and blank when it is running nothing, indicates that the GPU is idle 80\% of the time. Note, this baseline implementation is not intentionally hampered. Its code was introduced by ESPnet \cite{espnet-decoders} \cite{espnet}, and an adaptation has been used in NVIDIA NeMo \cite{nemo-decoders} \cite{nemo} for years.

The poor performance of the greedy decoder, despite being small, can be explained as ``kernel launch latency overhead", i.e., the GPU finishes executing each kernel long before the CPU requests the GPU to execute the next kernel.
CUDA Graphs addresses launch latency overhead by launching a group of kernels at once, to amortize individual launch cost. Unfortunately, prior versions of CUDA Graphs were not compatible with algorithms requiring data-dependent control flow, which includes RNN-T greedy search. CUDA 12.3 introduces conditional nodes to CUDA Graphs, which allow the GPU to handle the control flow autonomously \cite{cudagraphconditionalnodes}; both ``If" and ``While" nodes were added, though we use only ``While" nodes here.

\section{Related Work} \label{sec:relatedwork}

Some attempts to speed up RNN-T focus on changing the model architecture. Multi-blank transducers \cite{multiblank} introduce ``big blank" symbols to allow the decoder to skip frames of the encoder's output. Token-and-duration transducers (TDT) \cite{tdt} similarly allow for reducing the number of encoder frames processed, but decouple token prediction from duration prediction.

While these approaches focus on reducing the number of iterations, ours focuses on reducing the time taken by each iteration. Given that the GPU is idle 80\% of the time, our approach is orthogonal and can be easily combined with these other techniques. We demonstrate this for a TDT model in the results.

Naturally, there have been efforts to speed up the encoder of speech recognition models as well, e.g., ``fast conformer encoder" \cite{rekesh2023fast}, which all models evaluated in this paper use. Their techniques resulted in a significant reduction of encoder runtime, such that the decoder is becoming the bottleneck. Only 33\% of the runtime of the 1.1 billion parameter model is spend on the encoder, while the decoder consumers the remaining 67\% at batch size 32, c.f., Figure \ref{fig:encoder-vs-decoder}. Thus, the maximum speed-up we can achieve by further optimizing the encoder is $\frac{1}{1 - 0.33} = 1.49$, according to Amdahl's law. This theoretical result emphasizes the importance of improving decoding time.

There have also been efforts to speed up decoding by scheduling the algorithm's steps in a different way. For example, alignment-length sequence beam search decoding \cite{ALSD} and label looping greedy decoding \cite{LabelLooping} both seek to speed up decoding in batched scenarios, making decoding asynchronous among batch elements. This means that one element of a batch can be processing a different time step than another element of the same batch. This allows to make more efficient use of the GPU, as it reduces thread divergence. To illustrate, suppose that we ran at batch size 2 in Algorithm \ref{rnntgreedydecoding}, and set max\_symbols to 5. Suppose that the first batch element outputs 5 non-blanks every even time step, and one blank every odd time step. Suppose that the second batch element outputs 5 non-blanks every odd time step, and one blank every even time step. Algorithm \ref{rnntgreedydecoding} makes one element wait on the other element every ``outer loop" iteration; meanwhile, label looping allows for both elements to make continuous forward progress. The implementation methodology we describe here is compatible with these algorithms as well. We demonstrate how label looping greedy decoding and CUDA graphs complement each other in the results section.




\section{Algorithm Implementation}

The traditional way to submit work to a GPU is via CPU code launching a CUDA kernel one at a time through the ``cuLaunchKernel()" host function. To launch a GPU kernel, several things must be done in series: The CPU-side input parameters to the kernel are copied to a DMA region, the data is copied to the GPU,
and the GPU initializes its streaming multiprocessors to run the kernel. This process takes approximately 3 microseconds. The latency can be higher if the CPU has to do other work before launching a kernel, e.g., copying data back from the GPU. That can be the case for control flow, of which we have two instances, lines \ref{alg:outerloop} and \ref{alg:innerloop}, which we call the ``outer loop" and ``inner loop" in this paper. The outer loop runs ``max\_out\_len" times; this does not require a copy from the GPU to CPU except to initially compute ``max\_out\_len". However, for the inner loop, the CPU must copy ``not\_blank" from the GPU every iteration to evaluate the inner loop condition, since this value is computed every iteration; this can take over one microsecond. In total, the average time between kernel launches at batch size 32 for Parakeet RNN-T 1.1B is 5 microseconds, while the average kernel runtime is 1 microsecond, demonstrating that the GPU is idle 80\% of the time. In other words, if we could launch the decoder as a single CUDA graph, the decoder should run five times faster.

Normally, CUDA graphs are created in pytorch via ``CUDA stream capture to a graph". In that situation, CUDA kernels are launched to a dummy CUDA stream that does not actually run any kernels on the GPU. Rather, the launched kernels and their parameters are recorded, and they are later saved to a graph. There are a few challenges for our case:



\begin{enumerate}
\item Synchronization of the GPU to the host CPU is disallowed during stream capture.
\item CUDA stream capture to a graph cannot capture control flow nodes. Those must be inserted manually.
\end{enumerate}
We describe how we handle these issues for RNN-T greedy decoding in the following two sub-sections.

\subsection{Removal of host synchronization}

In order to capture a workload to a CUDA graph, it must not communicate with the host. There are a few surprising ways that this can happen in PyTorch. We document some here:
\begin{itemize}
\item If a torch.Tensor is in a context where it needs to be used on the CPU as a scalar (e.g., for control flow), the ``item()" method will be implicitly called on it to convert it to a Python Boolean on the host. This happens in line \ref{alg:innerloop}. We address this by doing control flow on the GPU; see Section \ref{sec:data-dependt-control-flow}.
\item If a, b, and mask are all torch.Tensors on the GPU, `a[mask] = b[mask]' requires synchronization with the host in eager mode PyTorch because space for the evaluated result of b[mask] must be calculated. This happens in lines \ref{alg:masking1} and \ref{alg:masking2}. This problem can be avoided by using this instead: `torch.where(mask, b, a, out=a)'.
\item Normally, indexing torch.Tensor \emph{a} with another torch.Tensor \emph{b} (\emph{a[b]}) does not synchronize with the host. However, if b is a scalar, the GPU does synchronize with the host, because of legacy behavior described in pytorch issue \#105641\footnote{\url{https://github.com/pytorch/pytorch/issues/105641}}. We originally encountered this issue in line \ref{alg:savekv}. Using torch.Tensor.index\_copy() can be used as a workaround.
\end{itemize}
There are other situations that can cause PyTorch code to synchronize with the host. We recommend using torch.cuda.set\_sync\_debug\_mode(2) to detect these.

\subsection{Inserting data-dependent control flow} \label{sec:data-dependt-control-flow}

We cannot use stream capture to capture control flow, because control flow is done on the CPU, not on the GPU. In order to insert a while loop into a CUDA graph, we need to (1) insert a CUDA kernel that evaluates the while loop's condition and then sets a global boolean to that value via the syscall cudaGraphSetConditional(), (2) insert a ``while type" conditional node that executes its ``subgraph" 
when that global boolean condition is true, and (3) finally capture the body of the while loop to this subgraph. The subgraph must also contain the kernel which calls cudaGraphSetConditional() mentioned in (1), to reset the while loop condition. Because the kernel that calls cudaGraphSetConditional() is inserted both before the while loop node and at the end of the while loop's subgraph, this mechanical process can be implemented with a Python context manager. We simply replace lines \ref{alg:outerloop} and \ref{alg:innerloop} with a Python ``with" statement calling this context manager. The exact implementation can be found at \url{https://github.com/NVIDIA/NeMo/blob/main/nemo/collections/asr/parts/submodules/cuda_graph_rnnt_greedy_decoding.py}.

\begin{figure}[t]
    \centering
    \includegraphics[width=1\linewidth]{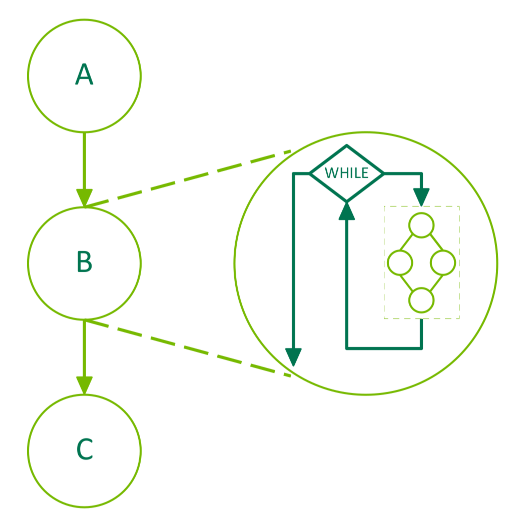}
    \caption{Schematic diagram of a CUDA graph with a while loop. The left hand side chain is a CUDA graph of nodes, whose B node is a while loop type ``conditional node." This node itself contains a ``child" CUDA graph within it.}
    \label{fig:cuda-graph-conditionals}
\end{figure}

Figure \ref{fig:cuda-graph-conditionals} depicts how this works. Node A is the kernel node which calls cudaGraphSetConditional(), node B is a conditional node that implements a while loop. Upon first entry, it checks its global boolean. If it is false, node C executes next. If it is true, node B's subgraph executes. After node B's subgraph executes, the global boolean is evaluated again to determine whether to run the loop body subgraph again or node C.



\begin{algorithm}
\caption{Greedy Decoding with RNN-T Model, with a one-layer RNN prediction network}
\begin{algorithmic}[1]
\Procedure{GreedyDecode}{$\text{x:float[B,T,F]}, \text{out\_len:int[B]},\text{max\_symbols: int}$}
    \State $\text{hidden = torch.zeros((B, RNN\_Hidden\_Dim))}$
    \State $\text{max\_out\_len = out\_len.max()}$
    \For{$\text{time\_idx in range(max\_out\_len)}$}
    \label{alg:outerloop}
        \State $\text{f} = \text{x}.\text{narrow}(\text{dim=1}, \text{start=time\_idx}, \text{length=1})$
        \State $\text{not\_blank = torch.tensor(True)}$
        \State $\text{symbols\_added = torch.tensor(0)}$
        \State $\text{blank\_mask = time\_idx} \geq \text{out\_len}$
        \While{$(\text{not\_blank}$ \textbf{and} $\text{symbols\_added} < \text{max\_symbols}).\text{item}()$}  \label{alg:innerloop}
            \State $\text{g, hidden\_prime = prediction(last\_label, hidden)}$
            \State $\text{logp = joint(f, g)}$
            \State $\text{k, v = logp.max()}$
            \State \textit{Save k and v for later} \label{alg:savekv}
            \State blank\_mask = blank\_mask $||$ (k == blank\_index)
            \State $\text{last\_label[\textbf{not} blank\_mask] = k[\textbf{not} blank\_mask]}$ 
            \label{alg:masking1} \State $\text{hidden[\textbf{not} blank\_mask] = hidden\_prime[\textbf{not} blank\_mask]}$ 
            \label{alg:masking2} \State $\text{not\_all\_blank\_t = (\textbf{not} blank\_mask).any()}$
            \State $\text{symbols\_added += 1}$
        \EndWhile
    \EndFor
    \State \textbf{return} $\textit{hypotheses built from saved values of k and v}$
\EndProcedure
\end{algorithmic}
\label{rnntgreedydecoding}
\end{algorithm}

\begin{table*}[!t]\centering
\caption{Word error rate and speed-up of tested models and decoding algorithms. The 
``Speed-up Factor" columns are computed relative to the first row of a given model's box (the ``Baseline" Decoder).}\label{tab:results}
\scriptsize
\begin{tabular}{lrrrrrr}\toprule
Model Name &Decoder &RTFx &\% Time in Decoder &WER &Overall Speed-up Factor &Decoder Speed-up Factor \\\midrule
Parakeet CTC 1.1B &Baseline &1336.24 &4.51 &3.78 &N/A &N/A \\
\hline
Parakeet RNN-T 1.1B &Baseline &414.63 &67.39 &2.7 &1 &1 \\
&CUDA Graphs &1040.23 &25.67 &2.7 &2.51 &6.59 \\
&Label Looping &659.72 &51.24 &2.7 &1.59 &2.09 \\
&Label Looping + CUDA Graphs &1120.88 &17.52 &2.7 &2.70 &10.40 \\
\hline
Parakeet TDT 1.1B &Label Looping &853.16 &36.80 &2.79 &1 &1 \\
&Label Looping + CUDA Graphs &1212.68 &10.57 &2.79 &1.42 &4.95 \\
\hline
& & & & & & \\
\hline
Parakeet CTC 0.6B &Baseline &2017.58 &8.46 &4.01 &N/A &N/A \\
\hline
Parakeet RNN-T 0.6B &Baseline &466.58 &75.30 &3.29 &1 &1 \\
&CUDA Graphs &1434.38 &34.61 &3.29 &3.07 &6.69 \\
&Label Looping &787.31 &62.31 &3.29 &1.69 &2.04 \\
&Label Looping + CUDA Graphs &1561.22 &24.745 &3.29 &3.35 &10.18 \\
\bottomrule
\end{tabular}
\end{table*}


\section{Results}

We benchmark our algorithms in a few scenarios in an offline inference task. For all results, we use an A100-80GiB PCIe card, and a batch size of 32, evaluated on Librispeech Test Other \cite{librispeech}, a 5.1 hour dataset. The dataset's utterances are sorted by duration, high to low. Bfloat16 is used for inference. The dataset is loaded with background dataloaders so that data loading time is not measured. max\_symbols is set to 5. Inference is run five times for warmup, and then ten times for measurements.

The usage of a relatively small batch size (32) deserves some explanation. Naturally, one would expect that larger batch sizes improve throughput until the system reaches 100\% utilization.
This holds true for non-autoregressive models like CTC, but not necessarily for autoregressive models. At the end of Section \ref{sec:relatedwork}, we describe a scenario where a particular combination of inputs runs particularly slow because they create
many tokens. Generally speaking for RNN-T decoding, only one or two symbols per outer loop time step are generated per time step, but an adversary can generate an input that always generates max\_symbols per outer loop time step. This will negatively impact throughput, as all elements in a batch must finish before any element's outputs is returned in current implementations. We can formalize this concept: Suppose the probability that a given input is adversarial is $p$. The probability that there is at least one adversarial(slow) input in a batch of size $k$ can be expressed as the cumulative distribution function of a geometric distribution: $P(X \le k) = 1 - (1 - p)^k$. We can see that the probability of at least one adversarial input rises exponentially with the batch size. Batch size 32 hits a sweet spot where we are able to utilize the GPU hardware fully, while also keeping the probability of being slowed down by a adversarial element low. 

We measure five models: Parakeet CTC 1.1B, Parakeet RNN-T 1.1B, Parakeet TDT 1.1B, Parakeet CTC 0.6B, and Parakeet RNN-T 0.6B. The approximate number of parameters in each model is given in each model's name. All models are publicly available on Huggingface \footnote{\url{https://huggingface.co/collections/nvidia/parakeet-659711f49d1469e51546e021}}.
There is no Parakeet TDT 0.6B. 

We investigated several algorithms:
\begin{enumerate}
    \item The baseline CTC greedy decoder (``Baseline").
    \item The baseline RNN-T greedy decoder (``Baseline").
    \item The baseline RNN-T greedy decoder re-implemented to be CUDA graph compatible (``CUDA Graphs").
    \item The label looping RNN-T greedy decoding algorithm without CUDA graphs (``Label Looping").
    \item The label looping RNN-T greedy decoding algorithm with CUDA graphs (``Label Looping + CUDA Graphs").
    \item The label looping TDT algorithm (``Label Looping").
    \item The baseline TDT algorithm sped up with CUDA Graphs (``Label Looping + CUDA Graphs").
\end{enumerate}

We do not use the baseline TDT algorithm because it is not exact at batch sizes greater than 1; see Section 5.2 of \cite{tdt}.
We use RTFx as the main measurement of performance. RTFx is the number of hours an ASR model can transcribe in one hour of wall clock time. We also measure word error rate (WER) to verify that our implementations produce exactly the same results as the reference (baseline) code.
While measuring ten inference runs we noticed that the standard deviation is at most 1\% from the average RTFx. Since there is so little run-to-run variance, we decided to omit confidence intervals from these measurements. Results are summarized in Table \ref{tab:results}. 
All optimized algorithms achieve the same WER as their respective baselines, proving that they are exact. ``Overall Speed-up Factor" refers to how much faster the end-to-end runtime is over the baseline, including encoder runtime. In the case of Parakeet TDT 1.1B, ``Label Looping" is considered the baseline. ``Decoder Speed-up Factor" refers to the speed up of just the decoder.

The CUDA graphs implementation of decoding for Parakeet RNN-T 1.1B is 6.59 times faster than the baseline implementation. This exceeds the theoretical 5 times speed-up from removing the 80\% idle time. The additional speed-up is caused by CUDA graphs eliminating GPU-side overheads beyond the scope of this paper. When cuda graphs are applied to label looping, decoding is 10.40 times faster than the baseline.


For overall (end-to-end) speeds, Parakeet RNN-T 1.1B runs 2.51 times faster than the baseline using just CUDA graphs, and 2.70 times faster combined with label looping. The TDT model is 1.42 times faster with CUDA graphs. This TDT model is only 9\% slower than the 1.1B parameter CTC model, which runs at 1336.24 RTFx. Meanwhile Parakeet RNN-T 1.1B with label looping and CUDA graphs decoding is only 16\% slower than this CTC model. Overall speeds are more pronounced for the 0.6B models because their encoders are smaller.



\section{Future Work}

Revisiting Amdahl's Law, for Parakeet RNN-T 1.1B, we have reduced the runtime consumed by the decoder from 70\% to 18\% by using labeling looping with cuda graphs. Therefore, speeding up the encoder is now the best course of action for further accelerating the end-to-end pipeline. In particular, weight-only quantization of Bfloat16 weights to int8/int4 weights can result in a speed-up the required matrix multiplications up to 35\%/56\% without accuracy loss in Transformer encoders \cite{RAWNHENRY}. 


Another avenue is to apply this work to other autoregressive models. Our results were achieved by using CUDA API calls manually, as described in Section \ref{sec:data-dependt-control-flow}. We are investigating how to bring conditional nodes directly into frameworks, e.g., via PyTorch's new higher-order ops like torch.while\_loop and torch.cond \cite{HigherOrderOpsPytorch}. This provides several benefits: it is easier for non-experts, compiler technologies like torch.compile can provide speed-ups from techniques like kernel fusion, and it can accelerate all autoregressive models, not just RNN-T. To give an example of potential benefit, we have observed in unpublished work that the GPU is idle 5\% of the time in an optimized Whisper inference implementation while it waits for the CPU to decide whether to call the autoregressive decoder again. While 5\% overhead is little, we expect this percentage to grow with future, faster GPU generations.



\section{Conclusion}

We demonstrated a new way for optimizing greedy decoding for RNN-T speech recognition models by reducing the latency of each iteration instead of reducing the number of iterations. It is complementary to existing efforts to reduce the total number of iterations, opens up new optimization possibilities by exposing unfused kernels in the decoder and the encoder, and generalizes to benefit any autoregressive model.

\newpage

\bibliographystyle{IEEEtran}
\bibliography{mybib}

\begin{thebibliography}{10}
\providecommand{\url}[1]{#1}
\csname url@samestyle\endcsname
\providecommand{\newblock}{\relax}
\providecommand{\bibinfo}[2]{#2}
\providecommand{\BIBentrySTDinterwordspacing}{\spaceskip=0pt\relax}
\providecommand{\BIBentryALTinterwordstretchfactor}{4}
\providecommand{\BIBentryALTinterwordspacing}{\spaceskip=\fontdimen2\font plus
\BIBentryALTinterwordstretchfactor\fontdimen3\font minus \fontdimen4\font\relax}
\providecommand{\BIBforeignlanguage}[2]{{%
\expandafter\ifx\csname l@#1\endcsname\relax
\typeout{** WARNING: IEEEtran.bst: No hyphenation pattern has been}%
\typeout{** loaded for the language `#1'. Using the pattern for}%
\typeout{** the default language instead.}%
\else
\language=\csname l@#1\endcsname
\fi
#2}}
\providecommand{\BIBdecl}{\relax}
\BIBdecl

\bibitem{rnnt}
\BIBentryALTinterwordspacing
A.~Graves, ``Sequence transduction with recurrent neural networks,'' \emph{CoRR}, vol. abs/1211.3711, 2012. [Online]. Available: \url{http://arxiv.org/abs/1211.3711}
\BIBentrySTDinterwordspacing

\bibitem{ctc}
\BIBentryALTinterwordspacing
A.~Graves, S.~Fern\'{a}ndez, F.~Gomez, and J.~Schmidhuber, ``Connectionist temporal classification: labelling unsegmented sequence data with recurrent neural networks,'' in \emph{Proceedings of the 23rd International Conference on Machine Learning}, ser. ICML '06.\hskip 1em plus 0.5em minus 0.4em\relax New York, NY, USA: Association for Computing Machinery, 2006, p. 369–376. [Online]. Available: \url{https://doi.org/10.1145/1143844.1143891}
\BIBentrySTDinterwordspacing

\bibitem{openasrleaderboard}
``Open asr leaderboard,'' \url{https://huggingface.co/spaces/hf-audio/open_asr_leaderboard}, 2023, accessed: 2024-02-29.

\bibitem{espnet-decoders}
``Transducer decoders in espnet,'' \url{https://github.com/espnet/espnet/blob/f8a6e763089c75963cd4071eda32ee36864cf5b8/espnet2/asr/transducer/beam_search_transducer.py}, 2023, accessed: 2024-02-29.

\bibitem{espnet}
\BIBentryALTinterwordspacing
S.~Watanabe, T.~Hori, S.~Karita, T.~Hayashi, J.~Nishitoba, Y.~Unno, N.~{Enrique Yalta Soplin}, J.~Heymann, M.~Wiesner, N.~Chen, A.~Renduchintala, and T.~Ochiai, ``{ESPnet}: End-to-end speech processing toolkit,'' in \emph{Proceedings of Interspeech}, 2018, pp. 2207--2211. [Online]. Available: \url{http://dx.doi.org/10.21437/Interspeech.2018-1456}
\BIBentrySTDinterwordspacing

\bibitem{nemo-decoders}
``Transducer decoders in nemo,'' \url{https://github.com/NVIDIA/NeMo/blob/d8ebaa5fb3b7df556b282f3547a605bded5b61de/nemo/collections/asr/parts/submodules/rnnt_decoding.py}, 2023, accessed: 2024-02-29.

\bibitem{nemo}
\BIBentryALTinterwordspacing
E.~Harper, S.~Majumdar, O.~Kuchaiev, L.~Jason, Y.~Zhang, E.~Bakhturina, V.~Noroozi, S.~Subramanian, K.~Nithin, H.~Jocelyn, F.~Jia, J.~Balam, X.~Yang, M.~Livne, Y.~Dong, S.~Naren, and B.~Ginsburg, ``{NeMo: a toolkit for Conversational AI and Large Language Models}.'' [Online]. Available: \url{https://github.com/NVIDIA/NeMo}
\BIBentrySTDinterwordspacing

\bibitem{cudagraphconditionalnodes}
``Conditional graph nodes,'' \url{https://docs.nvidia.com/cuda/cuda-c-programming-guide/index.html#conditional-graph-nodes}, 2024, accessed: 2024-03-07.

\bibitem{multiblank}
H.~Xu, F.~Jia, S.~Majumdar, S.~Watanabe, and B.~Ginsburg, ``Multi-blank transducers for speech recognition,'' \emph{arXiv:2211.03541}, 2022.

\bibitem{tdt}
H.~Xu, F.~Jia, S.~Majumdar, H.~Huang, S.~Watanabe, and B.~Ginsburg, ``Efficient sequence transduction by jointly predicting tokens and durations,'' \emph{arXiv preprint arXiv:2304.06795}, 2023.

\bibitem{rekesh2023fast}
D.~Rekesh, N.~R. Koluguri, S.~Kriman, S.~Majumdar, V.~Noroozi, H.~Huang, O.~Hrinchuk, K.~Puvvada, A.~Kumar, J.~Balam, and B.~Ginsburg, ``Fast conformer with linearly scalable attention for efficient speech recognition,'' 2023.

\bibitem{ALSD}
G.~Saon, Z.~T{\"u}ske, and K.~Audhkhasi, ``Alignment-length synchronous decoding for rnn transducer,'' in \emph{ICASSP 2020-2020 IEEE International Conference on Acoustics, Speech and Signal Processing (ICASSP)}.\hskip 1em plus 0.5em minus 0.4em\relax IEEE, 2020, pp. 7804--7808.

\bibitem{LabelLooping}
``Redacted because this paper is currently under review,'' 2024.

\bibitem{librispeech}
V.~Panayotov, G.~Chen, D.~Povey, and S.~Khudanpur, ``Librispeech: An asr corpus based on public domain audio books,'' in \emph{2015 IEEE International Conference on Acoustics, Speech and Signal Processing (ICASSP)}, 2015, pp. 5206--5210.

\bibitem{RAWNHENRY}
Y.~J. Kim, R.~Henry, R.~Fahim, and H.~H. Awadalla, ``Who says elephants can't run: Bringing large scale moe models into cloud scale production,'' 2022.

\bibitem{HigherOrderOpsPytorch}
{PyTorch Development Team}, ``Higher order ops in pytorch,'' \url{https://github.com/pytorch/pytorch/blob/main/torch/_inductor/cudagraph_trees.py}, 2023, accessed: 2024-02-29.

\end{thebibliography}

\end{document}